\documentclass[runningheads]{llncs}
\usepackage[T1]{fontenc}
\usepackage{graphicx}
\usepackage{hyperref}
\usepackage{subfigure}
\usepackage[misc,geometry]{ifsym}
\usepackage{color}

\begin{document}
\title{Few-shots Portrait Generation with Style Enhancement and Identity Preservation}
%
\titlerunning{Portrait Generation with StyleIdentityGAN}
%
\author{Runchuan Zhu\inst{2} 
\and
Naye Ji\inst{1,2}(\Letter)\orcidID{0000-0002-6986-3766} \and
Youbing Zhao\inst{1,2}
\and
Fan Zhang\inst{1,2}\orcidID{0000-0002-9534-1777}}
\authorrunning{Zhu et al.}
%
\institute{Institute}
\institute{College of Media Engineering, Communication University of Zhejiang, Hangzhou, China\and
Key Lab of Film and TV Media Technology of Zhejiang Province, Communication University of Zhejiang, Hangzhou, China\\
\email{277949825@qq.com}\\
\email{\{jinaye, zyb, fanzhang\}@cuz.edu.cn}}
\maketitle              
\begin{abstract}
Nowadays, the wide application of virtual digital human promotes the comprehensive prosperity and development of digital culture supported by digital economy. The personalized portrait automatically generated by AI technology needs both the natural artistic style and human sentiment. In this paper, we propose a novel StyleIdentityGAN model, which can ensure the identity and artistry of the generated portrait at the same time. Specifically, the style-enhanced module focuses on artistic style features decoupling and transferring to improve the artistry of generated virtual face images. Meanwhile, the identity-enhanced module preserves the significant features extracted from the input photo. Furthermore, the proposed method requires a small number of reference style data. Experiments demonstrate the superiority of StyleIdentityGAN over state-of-art methods in artistry and identity effects, with comparisons done qualitatively, quantitatively and through a perceptual user study. Code has been released on Github:\footnote{\href{https://github.com/Zrc007/StyleIdentityGAN} {https://github.com/Zrc007/StyleIdentityGAN}}.

\end{abstract}

\section{Introduction}
Stylization is the process which converts the style of the input image into one of the corresponding style reference samples. During the conversion process, the detailed features of the original input image should be kept as much as possible, while the line and texture of the reference are imitated. With the development of AI technology, image stylization can be widely used in video conferencing, entertainment and other fields in our daily life. Under the background of the massive demand for digital content in the current popular Meta Universe, the requirements of the automatic image stylization quality are becoming increasingly higher.

Stylization has a long history of research and can be divided into two categories: image-based style transfer and face-specific style transfer. Image-based style transfer is to render images with different styles and to retain the original image content as much as possible~\cite{jing2019neural}. However, traditional style transfer methods can only extract low-level features (color, texture, etc.) of images to synthesize textures without high-level features. In recent years, deep learning has developed rapidly and has been widely applied in many fields. Deep neural network has incisive feature extraction ability to obtain rich semantic information. Thanks to this, style transfer based on deep learning has made a series of progress.

In contrast, image-based style transfer is simpler than face-specific style transfer because the latter needs to pay more attention to the refinement of detail features and the integration of style. Face-specific style commonly includes sketch stylization, cartoon stylization and oil painting stylization from literature. In sketch stylization, facial feature information is mainly retained, while irrelevant colors are removed or converted into gray values; In cartoon stylization and oil painting stylization, since the result is considerably different from the original portrait, we need to preserve the original facial feature information and learn the color and texture. What is more, the scarcity of data is one of the challenges for stylization while many stylization algorithms are exceedingly dependent on dataset. When we encounter interesting style references that have not been incorporated into the dataset, we need new methods for stylization.

In this paper, we propose StyleIdentityGAN to automatically generate portraits of specified artistic style. The proposed StyleIdentityGAN model can ensure the identity and artistry of the generated portrait at the same time. The proposed method requires only a few reference style samples. Then, the style-enhanced module can decoupling artistic style features, such as shading tone, color, texture features from reference style samples. Based on the decoupled artistic style features that corresponds to the latest space, the artistic style features can be transferred to the required results, which can improve the artistry of generated portraits. After the artistry can be guaranteed, our identity-enhanced module seeks to preserve the identity information of the original face. We introduce a feature loss function to preserves the significant features of the input photo to guarantee identity characteristics. We conduct our experiments on 4 datasets including cartoon and sketch style. The qualitative and quantitative evaluation proved that the portrait results of our proposed StyleIdentityGAN exhibits good aesthetics and assimilate with the original face. Our contributions can be summarized as follows:
\begin{enumerate}
\item [(1)] 
We decouple the stylized texture features from stylized image domain to obtain the art style independent features. The decoupled style features can be obtained through transformation of the specified style latent space to enhance 
artistry.
\item [(2)] 
The introduced feature loss emphasizes the importance of facial features, which solves the problem that the output results are easy to lose the facial features of the input image while ensuring the stylized effect.
\item [(3)] 
Our few-shot strategy does not rely on large-scale data sets, which only needs a few pictures to start training. Moreover, the training is simple and fast. When generating a group of results with excellent effect, only a few minutes are needed.
\end{enumerate}

\section{Related Work}
\subsection{Image-based Style Transfer}
 Image-based style transfer can be divided into two groups: \textbf{(1) Image based iteration}: Gatys et al. proposed the most primitive style transfer algorithm~\cite{Gatys2016A}, which uses convolutional neural network to extract features, followed by texture synthesis, calculation of content loss and style loss, gradient descent to optimize the total loss, and constantly iterating the image to generate an artistic image. Afterwards, Liao proposed visual attribute transfer~\cite{liao2017visual} by combining deep VGG19 and patchmatch and achieved good results. \textbf{(2) Model based iteration}: One of them is based on the back propagation stylized model, which is mainly optimized in the model architecture. The other is based on GAN (generative adversarial networks). Johnson firstly proposed a real-time style transfer~\cite{johnson2016perceptual} based on iterative optimization generation model, which uses perceptual loss function to train the generation model for a specific style, providing a good idea for improving the efficiency of image style transfer. Zhu et al. proposed CycleGAN~\cite{Zhu2017Unpaired}, which eliminates the need for specific image pairs when converting images between different fields, and solves the problem of difficult pair data collection. Moreover, UNIT~\cite{liu2017unsupervised} which combines GAN and VAE, suites for the mutual conversion between real images and StarGAN realizes the conversion of multiple fields through one model.

\subsection{Face-specific Style Transfer}
Face-specific style transfer based on GAN model have emerged enormously as the promotion of image-based style transfer and the gradual enrichment of face stylization database resources. For example, on the APDrawing line drawing portrait dataset, there are APDrawingGAN~\cite{yi2019apdrawinggan} integrated with global and local generators, asymmetric cycle-structure GAN, and U2-Net~\cite{qin2020u2} generating portraits with plenty of details. On the WebCaricature dataset~\cite{huo2017webcaricature}, the CariGANs~\cite{cao2018carigans}, a network integrating CarigeoGAN, and CaristyGAN, with the semantic-CariGANs of learning comics through semantic shape transformation are proposed.

In recent years, more and more research work specific to cartoon and cartoon style face migration have emerged,  representatives include white box cartoon feature representation combined with VGG and GAN and U-GAT-IT~\cite{kim2019u} introducing attention mechanism and adopting AdaLIN. Based on the U-GAT-IT, Zhuang and Yang~\cite{zhuang2021few} use Soft-AdaLIN to transfer information on unpaired data to generate more refined animation faces. In addition, based on the pre-training model of stylegan2~\cite{karras2020analyzing}, AgileGAN~\cite{song2021agilegan} applies enormous samples for fine tuning to achieve the generation results of comics and other styles, the artistry of which has been improved compared with the previous method. However, due to the small-scale dataset, the current stylization still has potential for improvement. In order to solve the problem, some one-shot characterized methods are proposed, such as DiFa~\cite{yu2019free} and JoJoGAN~\cite{chong2021jojogan}.

\subsection{Face Editing}
Face editing aims to manipulate certain attributes of the face image to generate series of new faces with the required attributes while preserving other details. Face editing is always based on face generation. Although using face key points to control faces can also realize face attribute editing, that often leads to face distortion and asymmetry. With the development of generation model, face editing has made profound progress. GAN realizes the mapping of images from one domain to another, which brings the emergence of high-quality papers on face editing. For example, AttGAN~\cite{he2019attgan} introduces attributive classification constraint, which could protect other non-edited attributes while ensuring the high quality of the generated image. SGGAN~\cite{wang2019semi} proposes a novel piece-wise guided Generative Adversarial Network, which uses semantic segmentation to further improve the generation performance and provide spatial mapping. SC-FEGAN~\cite{jo2019sc} takes another step in face editing, and realizes free face editing through the common input network of free-form original pictures, sketches, mask images, color images and noise. Face editing realized in this way will no longer be limited to attribute tags, but hand over the freedom of face editing to users, and realize face editing through joint input and SN-PatchGAN~\cite{yu2019free}. PSGAN~\cite{jiang2020psgan} proposes MDNet and AMM modules to migrate the makeup on any reference image to the source image without makeup. Jin et al.~\cite{jin2022attribute} proposed a method by aesthetics driven reinforcement learning.

\section{StyleIdentityGAN}

\subsection{Problem Definition}
\begin{figure}[ht]
\centering
\includegraphics[scale=0.34]{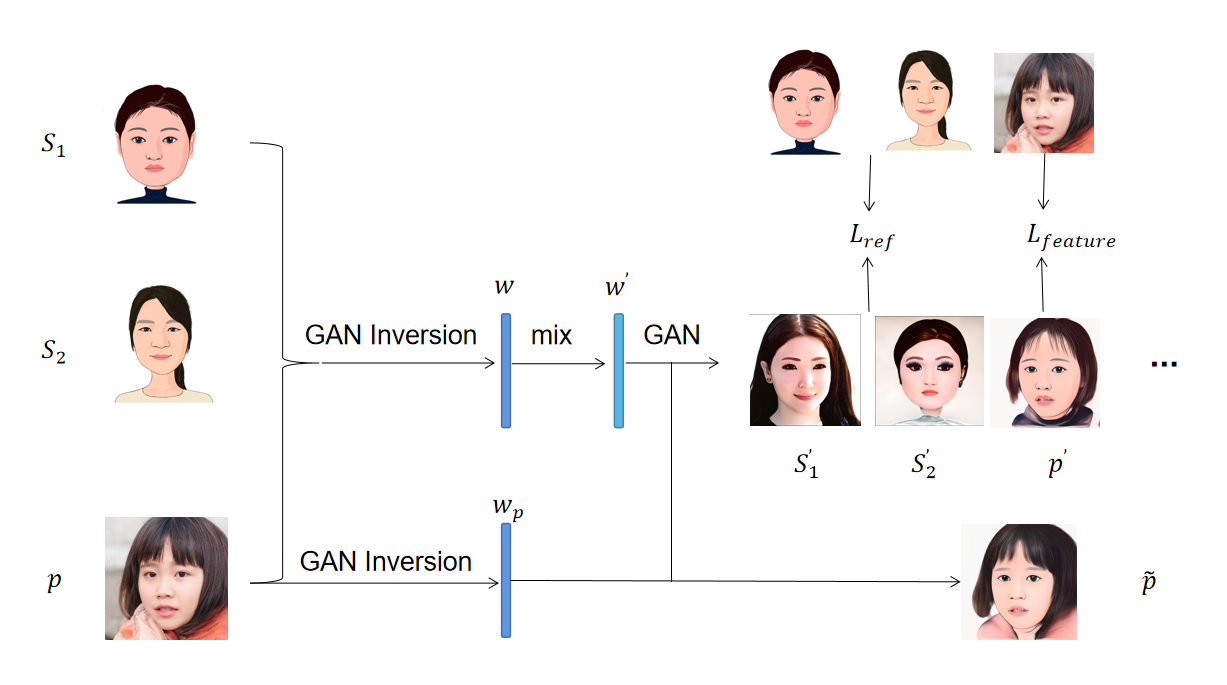}
\caption{Overview of our StyleIdentityGAN}
\label{fig:overview}
\end{figure}

Suppose that $\mathcal P$ is the face image domain and $\mathcal S$ is the stylized image domain. For any input face image $p \in  \mathcal P$, the corresponding stylized result is $\tilde{p} \cong s, s \in \mathcal S$, where $s$ is the specified artistic style sample image to be migrated. The goal of stylized portrait generation, which takes into account the complex structure and artistry of faces, is to build a mapping $G$ from the face image domain to the stylized image domain. Consequently, the generated $\tilde{p}$ still retains the identity information of the original human face $p$, that is, the structural and semantic features of the original face remain unchanged, but the style is as close as possible to the given sample $s$. The overview of our StyleIdentityGAN model is shown in Fig.~\ref{fig:overview}. The style-enhanced module and identity-enhanced module are described in Sec~\ref{sec32} and Sec~\ref{sec33}, respectively.

\subsection{Style-enhanced Module}\label{sec32}
We first decompose the stylized texture features from stylized image domain $\mathcal{S}$. Suppose the feature of the designated artistic reference is $\omega_s$, and then the style-enhanced module decouples the artistic style related features and artistic style independent features according to the stylization domain $\mathcal{S}$. Define the art style related feature as $\omega^{re}_s$, and the art style independent feature as $\omega^{un}_s$. After decoupling operation $F$ to $\omega_s$, the art style independent feature $\omega^{re}_s$ can be obtained:$\omega^{re}_s = \omega_s - \omega^{un}_s$. Because it depends on the style field $\mathcal{S}$, the $\omega^{re}_s$ of the specified style can be obtained from the field transformation $T$. The style features enhanced by artistic features are obtained through transformation.

We consider using few-shot strategy to eliminate the impact of single reference on the results on contrast with JoJoGAN's~\cite{chong2021jojogan} one-shot strategy. 
Compared with single reference, few-shot is easier to eliminate some effects of feature on the results.
Given an input face photo $p$ and a small group of reference sample images $s_1,...,s_n$, 
we deduce the corresponding latest space $\omega$ by using StyleGAN inversion for $p$ and $s_i$, respectively. The latent space $\omega$ has the function of encoding the content code of the main semantics in the image. Next, we randomly generate a tensor $\omega^{re}_{rand}$ with the same size as $\omega^{re}_s$, and then the new $\omega_s$ is:
\begin{equation}\label{eq1}
    \omega_s = \omega^{un}_s + \alpha * \omega^{re}_s + (1 - \alpha) * \omega^{re}_{rand}
\end{equation}
Acoording to Eq.~\ref{eq1}, we can generate new images $S_1^{'},...,S_n^{'}$ and $p^{'}$ by passing the new $\omega$ through a network $G$, which is originally StyleGAN. These images still retain the facial features of the original image, but change the image style, as shown in Figure. Finally, our destination is to obtain a network that can only change style without changing facial feature, so we loop steps 2 and 3 to train net $G$.

\subsection{Identity-enhanced Module}\label{sec33}
We notice that during the training process, the skin color and facial features of the generated image will gradually become close to the reference, so we hope to take some actions to suppress this change. The method we use is to add the feature loss, which is the Learned Perceptual Image Patch Similarity(LPIPS) of the current generated image and the input, and we record it as $\mathcal L_{feature}$, the total loss function is:
\begin{equation}
        \mathcal L = \mathcal L_{ref} + \lambda_{feature} * \mathcal L_{feature}
\end{equation}

\section{Experiments}
\subsection{Datasets} 
We conduct experiments on 4 datasets in sketch stylization and cartoon stylization. For shading style sketch stylization, we use CUFS~\cite{wang2009faceCUFS}, which includes 606 photo-sketch pairs, is used to study face sketch synthesis and face sketch recognition. Another dataset is a line style sketch dataset APDrawing~\cite{yi2019apdrawinggan}, which collects 140 pairs of facial photos and corresponding portraits. Besides, the cartoon stylization includes a dataset CAS-WACO collected by us of 50 face images with corresponding watercolor cartoon portraits drawn by professional artists, as well as a cartoon dataset of 317 cartoon face images from Toonify~\cite{Justin2020Tonnify}.

\subsection{Implementation details}

During the training, we set $\omega_{rand}$ and its hyperparameter $\alpha$ to 0.5. And we will adjust the $\lambda_{feature}$, within the range of $[0.0005,0.003]$, because we learned through a large number of experiments that lambda has the best generation effect in this range. According to experimental experience, some components of $\omega_s$ will have a greater impact on style, so we define an array swap$\_$list=[7, 9, 11, 15, 16, 17], during the  $\omega^{re}_{rand}$ generation, we do not change the values of other components, but only randomly generate swap$\_$list value of vector. In sketch stylization, we will set epoch to $150$, while the epoch of cartoon stylization is $500$. The epoch of sketch stylization is relatively small because we found in the experiment that the loss function has converged and the image has a good result at about $100$ epoch, and the subsequent iteration has not significantly improved the result. We used Tesla P40 (24GB) GPU for training and each reference takes about 30 seconds to sketch stylization, but $2$ minutes to cartoon stylization.

\subsection{Ablation Study}
\begin{figure}[ht]
\centering
\includegraphics[scale=0.3]{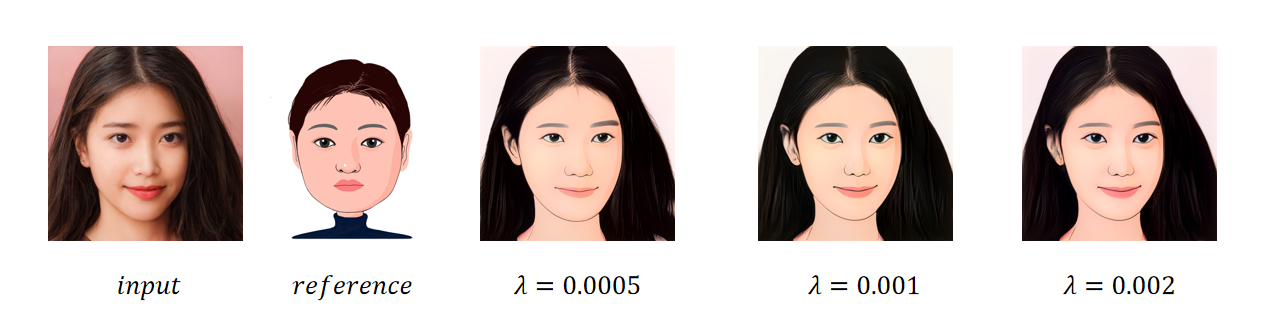}
\caption{Comparison of different $\lambda_{feature}$ in cartoon stylizaiton results on the CAS-WACO dataset.}
\label{fig:ablation}
\end{figure}

\begin{figure}[ht]
\centering
\includegraphics[scale=0.3]{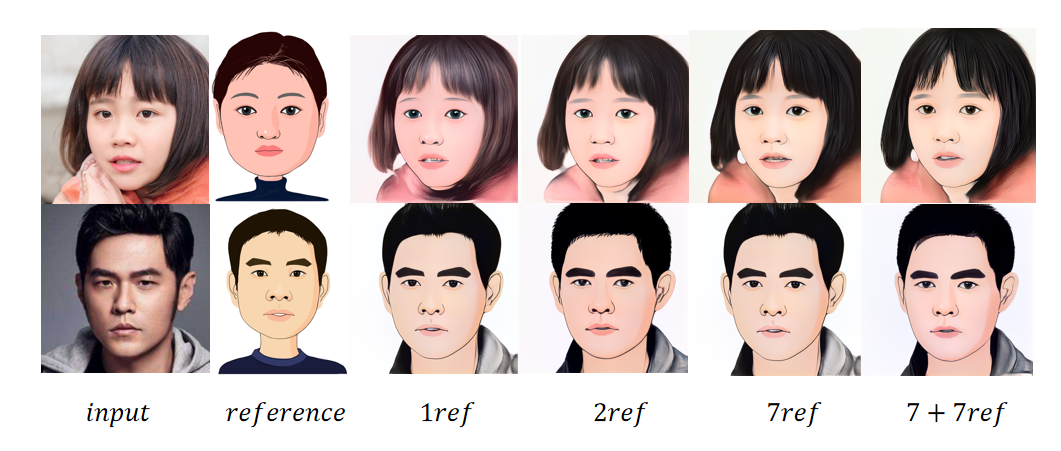}
\caption{Comparison of different reference number in cartoon stylizaiton results on the CAS-WACO dataset.}
\label{fig:ablation2}
\end{figure}

In the ablation study, we selected CAS-WACO dataset. Through a large number of experiments, we found that $\lambda_{feature}$ has a better effect in $[0.0005, 0.003]$. Therefore, we selected lambda values of $0.0005, 0.001, 0.002$, and Fig.~\ref{fig:ablation} as the experimental results in this experiment. We found that when $\lambda_{feature}$ is $0.001$, the resulting graph most meets our requirements. Therefore, we set $\lambda_{feature}$ values to $1$. Fig.~\ref{fig:ablation2} shows the impact of different reference numbers on the results in the following experiments. We selected 1, 2, 7, and 14 style maps respectively, of which 1, 2, and 7 were the same style maps as the input map, and 14 were 7 men and 7 women. Through the experimental results, we found that few shot would be more universal than one shot, and the results generated by one shot would be greatly affected by the facial features of the style map, which is not desired.

\subsection{Comparable Experiments}
\subsubsection{Qualitative Results.}
\begin{figure}[ht]
\centering
\includegraphics[scale=0.4]{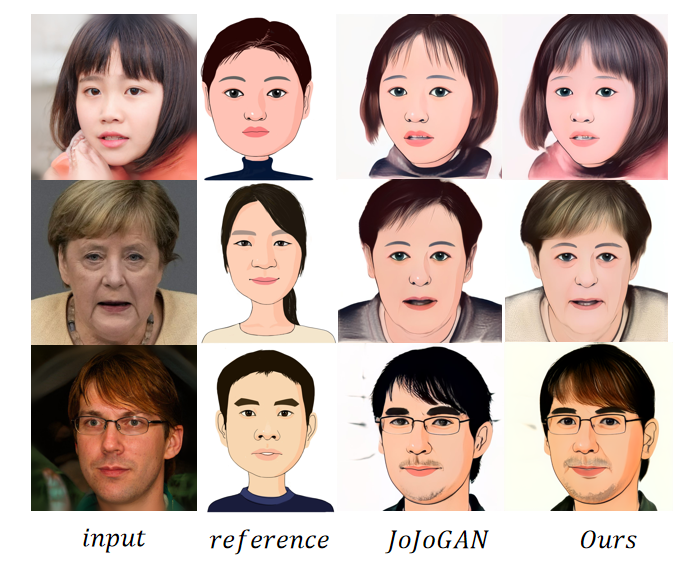}
\caption{Comparison of our method with JoJoGAN in CAS-WACO cartoon dataset.}
\label{fig:compare2}
\end{figure}

We compare our method with JoJoGAN~\cite{chong2021jojogan} in watercolor cartoon style, and APDrawingGAN~\cite{yi2019apdrawinggan}, U2Net~\cite{qin2020u2} in line sketch style. Due to convergence difficulties and GPU memory limitations, those methods were not able to directly support 1024×1024 resolution, thus we kept their original sizes of 256×256 or 512×512 for training and up-sampled the output to 1024×1024 for comparison. From Fig.~\ref{fig:compare2}, it can be seen that our method successfully cartoonized subjects with visually pleasing results. 

\begin{figure}[ht]
\centering
\includegraphics[scale=0.32]{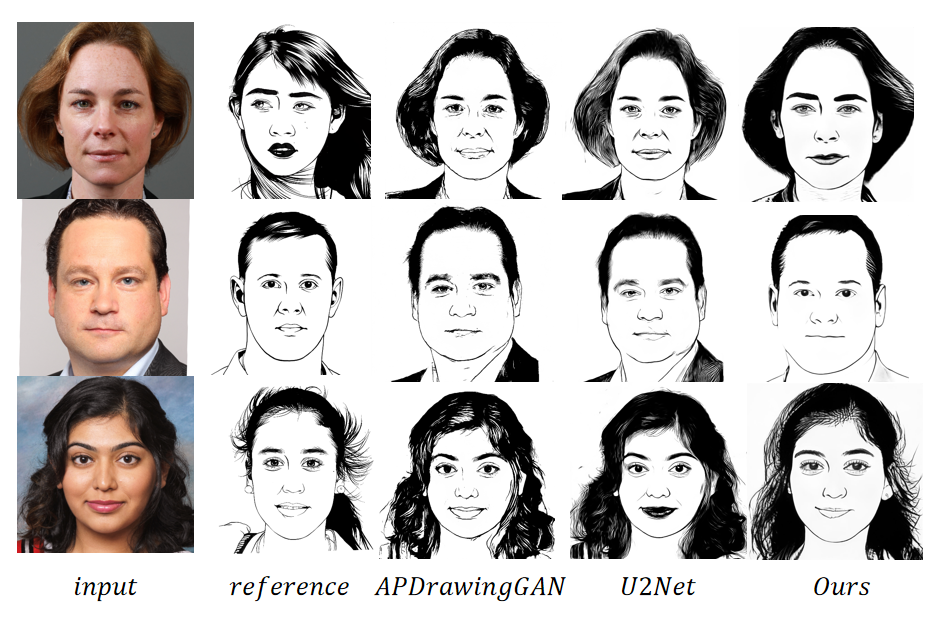}
\caption{Comparison of our method with U2Net and APDrawingGAN in APDrawing sketch dataset.}
\label{fig:compare1}
\end{figure}

\subsubsection{User Study.}
We further quantify improvement in stylization quality to try different parameters through human evaluation. We conducted a perceptual user study in which 28 participants were shown stylization results from different methods and different parameters. Among them, 11 participants are good at portrait drawing, while 17 participants are not skilled in art drawing. They were asked to select the best cartoonized images. Each participant was firstly shown 11 photo-portrait pairs of watercolor cartoon style of different methods. 
Table~\ref{tab:user} shows that results from our proposed method had the majority preference. Besides, the few-shot strategy outperforms one-shot strategy of JoJoGAN and our StyleIdentityGAN.
\begin{table}[]
\centering
\caption{User Study of watercolor cartoon stylization on CAS-WACO dataset}
\label{tab:user}
\begin{tabular}{c|cc|cc}
\hline
Method &\multicolumn{2}{c|}{JoJoGAN} & \multicolumn{2}{c}{Ours}  \\ \hline
Strategy  &\multicolumn{1}{c|}{One-shot}& \multicolumn{1}{c|}{Few-shot} & \multicolumn{1}{c|}{One-shot}  & {Few-shot} \\ \hline
Preference Score\% ↑ &\multicolumn{1}{c|}{3.00} & \multicolumn{1}{c|}{14.00} & \multicolumn{1}{c|}{29.22} &{48.74}\\\hline
\end{tabular}
\end{table}

\begin{figure}[ht]
\centering
\includegraphics[scale=0.42]{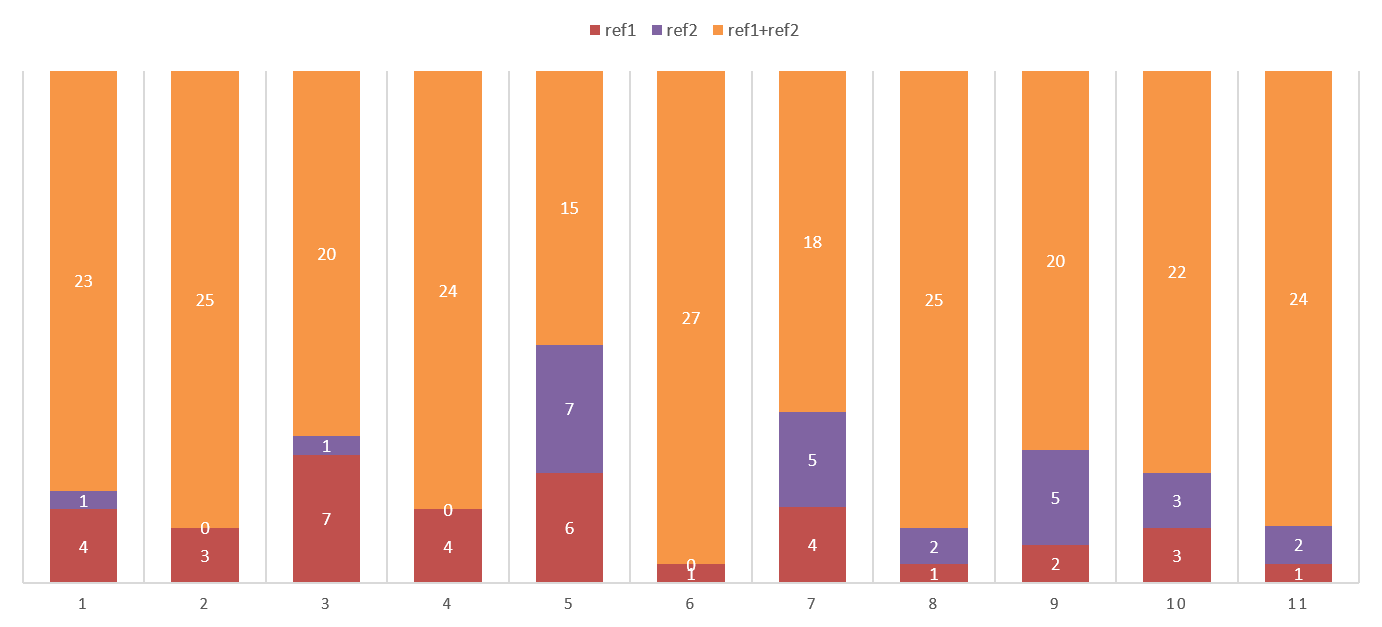}
\caption{Comparison of few-shot strategy with one-shot strategy of our method. The user study is tested on Tonnify cartoon style photo-portrait pairs.}
\label{fig:cartoonuser}
\end{figure}

Then each participant was shown 11 photo-portrait pairs of Tonnify cartoon style of different strategies. The human evaluation results is shown in Fig.~\ref{fig:cartoonuser}. Fig.~\ref{fig:cartoonuser} illustrates that portraits generated by one shot strategy of either style received fewer votes than portraits generated by few-shot strategy. It can be intuitively seen that our few-shot strategy has significant improvement over one-shot strategy on cartoon generation of Tonnify dataset.
 
\subsubsection{Quantification Evaluation.}
We choose the Fréchet Inception Distance (FID) score~\cite{Heusel2017FID} to quantitatively evaluate the stylization results. FID measures the visual similarity and distribution between two datasets of images. Each method generated stylized images from the CelebA-HQ dataset as input, and we computed FID to train the cartoon dataset. We can see from Table~\ref{tab:fid} that our method also performed best on this metric compared with JoJoGAN~\cite{chong2021jojogan} in watercolor cartoon style. 
Although it should be noted that since there are fewer than 5K images in the CelebA-HQ test set, FID scores may not be very reliable. Our results is quite equivalent to ones of APDrawingGAN~\cite{yi2019apdrawinggan} and U2Net~\cite{qin2020u2} in line sketch style. Meanwhile, we need few reference images and short time.

Another metric to quantitatively evaluate generative quality is the Structural Similarity (SSIM) measure~\cite{Wang2014SSIM}. SSIM is a widely used metric which computes structure similarity, luminance and contrast comparison using a sliding window on the local patch.

\begin{table}[]
\centering
\caption{FID and SSIM results}
\label{tab:fid}
\begin{tabular}{c|ccc|cc}
\hline
Dataset &\multicolumn{2}{c|}{CAS-WACO} & \multicolumn{3}{c}{APDrawing}  \\ \hline
Method  &\multicolumn{1}{c|}{JoJoGAN}& \multicolumn{1}{c|}{Ours} &           \multicolumn{1}{c|}{APDrawingGAN} &\multicolumn{1}{c|}{U2Net}           & {Ours}\\ \hline
FID ↓   &\multicolumn{1}{c|}{272.94} & \multicolumn{1}{c|}{\textbf{210.98}} &\multicolumn{1}{c|}{217.86}      &\multicolumn{1}{c|}{\textbf{208.74}}&{221.38}\\\hline
SSIM ↑  &\multicolumn{1}{c|}{\textbf{0.42}}   & \multicolumn{1}{c|}{0.40}   &\multicolumn{1}{c|}{0.43}       &\multicolumn{1}{c|}{\textbf{0.51}}  &{0.46}\\  \hline
\end{tabular}
\end{table}

\subsection{More Results and Discussion}
\begin{figure}[htbp]
\centering
\includegraphics[scale=0.35]{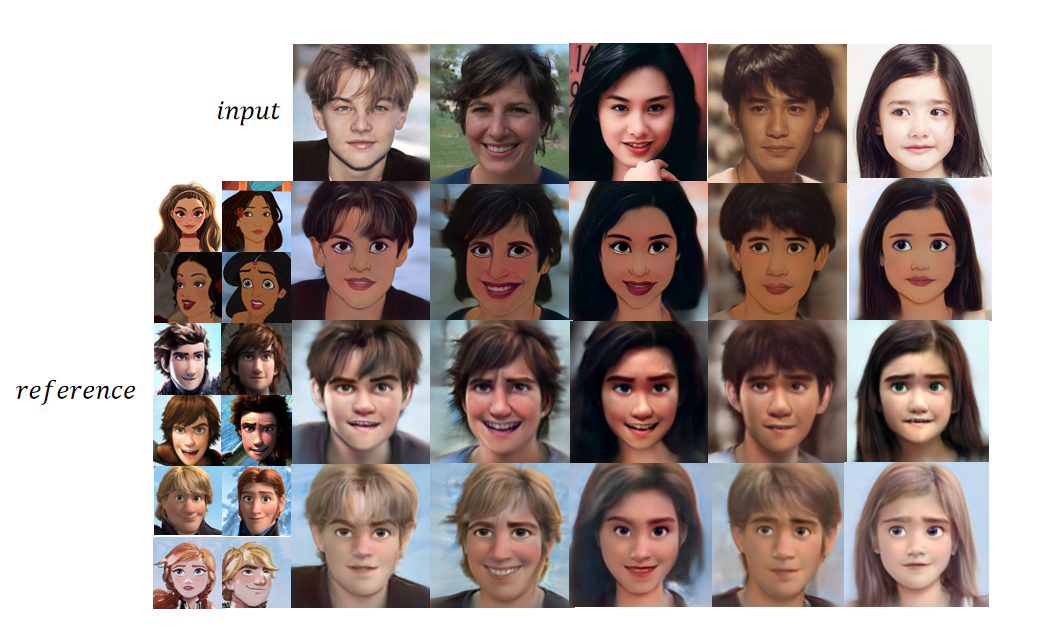}
\caption{Results of our method on Tonnify cartoon dataset.}
\label{fig:tonnifyresult}
\end{figure}

Fig.~\ref{fig:tonnifyresult} shows some results of our method on CUFS sketch dataset. Fig.~\ref{fig:tonnifyresult} shows results of our method using 4 reference style samples on Tonnify cartoon dataset. The various styles of Tonnify cartoon dataset cause artifacts on some local face components, e.g. ghosting on noses if using one-shot strategy. More comparison results are attached in the supplemental material. 

From the above results and the qualitative and quantitative evaluations, we can see the advantages of few-shot over one-shot. The results obtained by few-shot are more natural and do not depend heavily on the reference. In sketch stylization, compared with APDrawingGAN and U2Net, our method has this equivalent evaluation result, but our method only requires few references and has extremely short training time, which is our advantage. In cartoon stylization, compared with JoJoGAN, our results will be better in both quantitative evaluation and subjective evaluation, and we retain more facial features.

\begin{figure}[htbp]
\centering
\includegraphics[scale=0.3]{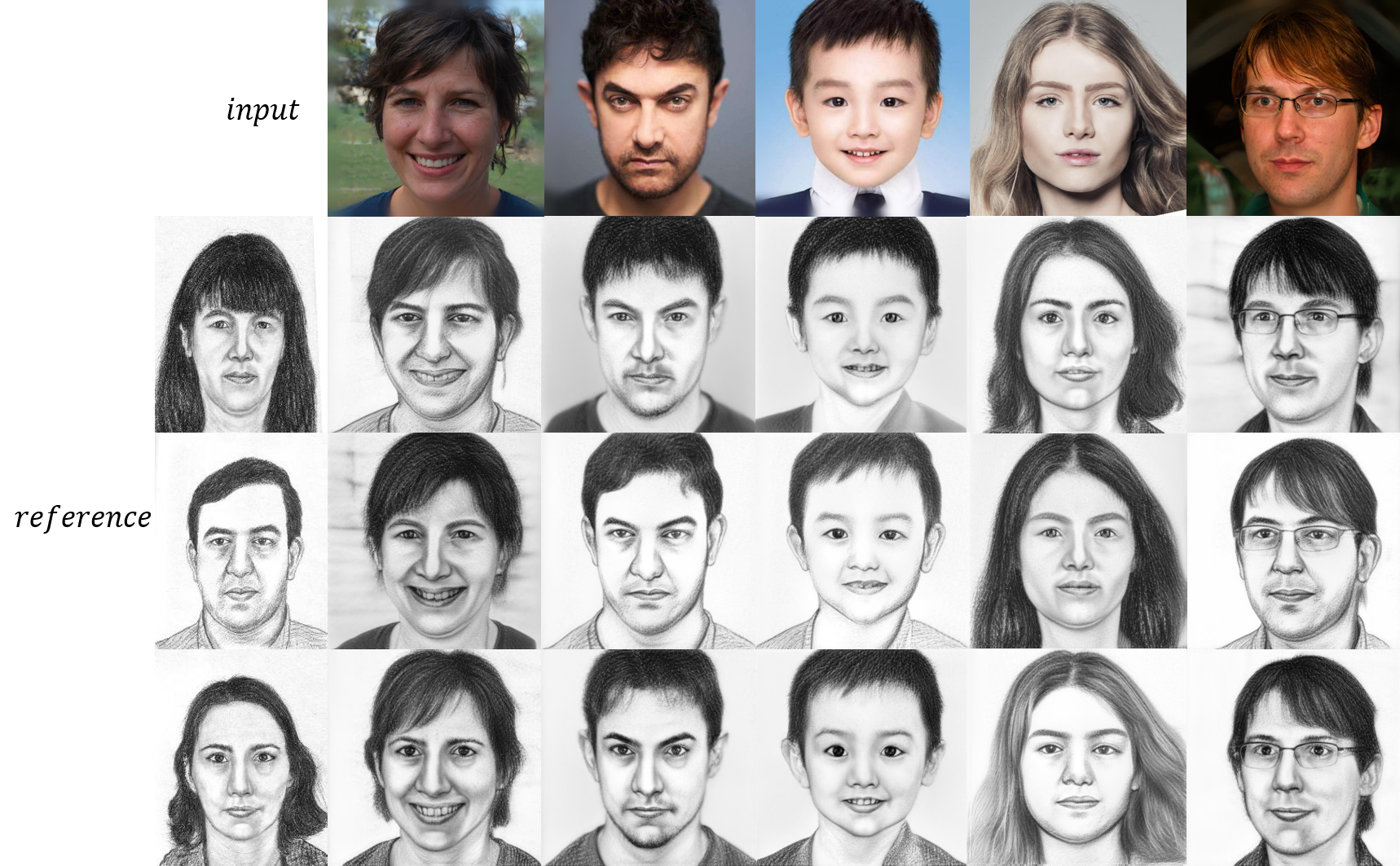}
\caption{Results of our method on CUFS sketch dataset.}
\label{fig:cufsresult}
\end{figure}

\section{Conclusion and Future Work}
In this paper, we propose a StyleIdentityGAN model which consists of a style-enhanced module and an identity-enhanced module. The former solves the problem of over dependence of the generated results on one-shot, and improves the artistic perception from a few reference style samples, while the latter reasonably retains the facial features of the input. In the future work, we will try to focus on optimizing facial details, e.g. hair naturalness, attention mechanism or adding normalizing flows to the style-enhanced module. On the basis of the preliminary improvement of artistic effect, emotional semantic features can be added to the structural semantic space in the generation process, and the facial structural features are fine tuned, which can edit and change the expression of the generated virtual face, making the generated virtual person more emotional and vivid. Furthermore, a model combined with implicit emotion and pose features can be built to generate video-driven virtual facial animations.
%
%
%

\bibliographystyle{splncs04}
\bibliography{mybibliography}

\begin{thebibliography}{10}
\providecommand{\url}[1]{\texttt{#1}}
\providecommand{\urlprefix}{URL }
\providecommand{\doi}[1]{https://doi.org/#1}

\bibitem{cao2018carigans}
Cao, K., Liao, J., Yuan, L.: Carigans: Unpaired photo-to-caricature
  translation. ACM Transactions on Graphics (TOG)  \textbf{37}(6),
  244.1--244.14 (2018)

\bibitem{chong2021jojogan}
Chong, M.J., Forsyth, D.: Jojogan: One shot face stylization. arXiv preprint
  arXiv:2112.11641  (2021)

\bibitem{Gatys2016A}
Gatys, L., Ecker, A., Bethge, M.: A neural algorithm of artistic style. Journal
  of Vision  \textbf{16}(12),  326--326 (2016),
  \url{https://doi.org/10.1167/16.12.326}

\bibitem{he2019attgan}
He, Z., Zuo, W., Kan, M., Shan, S., Chen, X.: Attgan: Facial attribute editing
  by only changing what you want. IEEE Transactions on Pattern Analysis and
  Machine Intelligence (PAMI)  \textbf{28}(11),  5464--5478 (2019)

\bibitem{Heusel2017FID}
Heusel, M., Ramsauer, H., Unterthiner, T., Nessler, B., Hochreiter, S.: Gans
  trained by a two time-scale update rule converge to a local nash equilibrium.
  In: Proceedings of the 31st International Conference on Neural Information
  Processing Systems. p. 6629–6640. NIPS'17, Curran Associates Inc., Red
  Hook, NY, USA (2017)

\bibitem{huo2017webcaricature}
Huo, J., Li, W., Shi, Y., Gao, Y., Yin, H.: Webcaricature: a benchmark for
  caricature recognition. In: British Machine Vision Conference (BMVC) (2017)

\bibitem{jiang2020psgan}
Jiang, W., Liu, S., Gao, C., Cao, J., He, R., Feng, J., Yan, S.: Psgan: Pose
  and expression robust spatial-aware gan for customizable makeup transfer. In:
  Proceedings of the IEEE/CVF Conference on Computer Vision and Pattern
  Recognition (CVPR). pp. 5194--5202 (2020)

\bibitem{jin2022attribute}
Jin, X., Zhao, S., Zhang, L., Zhao, X., Deng, Q., Xiao, C.: Attribute
  controllable beautiful caucasian face generation by aesthetics driven
  reinforcement learning. In: ACM Multimedia, Technical Demos and Videos
  Program (2022)

\bibitem{jing2019neural}
Jing, Y., Yang, Y., Feng, Z., Ye, J., Yu, Y., Song, M.: Neural style transfer:
  A review. IEEE Transactions on Visualization and Computer Graphics (TVCG)
  \textbf{26}(11),  3365--3385 (2019)

\bibitem{jo2019sc}
Jo, Y., Park, J.: Sc-fegan: Face editing generative adversarial network with
  user's sketch and color. In: Proceedings of the IEEE/CVF International
  Conference on Computer Vision (CVPR). pp. 1745--1753 (2019)

\bibitem{johnson2016perceptual}
Johnson, J., Alahi, A., Fei-Fei, L.: Perceptual losses for real-time style
  transfer and super-resolution. In: European Conference on Computer Vision
  (ECCV). pp. 694--711. Springer (2016)

\bibitem{karras2020analyzing}
Karras, T., Laine, S., Aittala, M., Hellsten, J., Lehtinen, J., Aila, T.:
  Analyzing and improving the image quality of stylegan. In: Proceedings of the
  IEEE/CVF International Conference on Computer Vision (CVPR). pp. 8110--8119
  (2020)

\bibitem{kim2019u}
Kim, J., Kim, M., Kang, H., Lee, K.: U-gat-it: Unsupervised generative
  attentional networks with adaptive layer-instance normalization for
  image-to-image translation. In: International Conference on Learning
  Representations (ICLR) (2020)

\bibitem{liao2017visual}
Liao, J., Yao, Y., Yuan, L., Hua, G., Kang, S.B.: Visual attribute transfer
  through deep image analogy. ACM Transactions on Graphics (TOG)
  \textbf{36}(4),  Article 120 (2017). \doi{10.1145/3072959.3073683}

\bibitem{liu2017unsupervised}
Liu, M.Y., Breuel, T., Kautz, J.: Unsupervised image-to-image translation
  networks. Advances in Neural Information Processing Systems  \textbf{30}
  (2017)

\bibitem{Justin2020Tonnify}
Pinkney, J.N., Adler, D.: Resolution dependent gan interpolation for
  controllable image synthesis between domains. arXiv preprint arXiv:2010.05334
   (2020)

\bibitem{qin2020u2}
Qin, X., Zhang, Z., Huang, C., Dehghan, M., Zaiane, O.R., Jagersand, M.:
  U2-net: Going deeper with nested u-structure for salient object detection.
  Pattern Recognition  \textbf{106},  107404 (2020)

\bibitem{song2021agilegan}
Song, G., Luo, L., Liu, J., Ma, W.C., Lai, C., Zheng, C., Cham, T.J.: Agilegan:
  stylizing portraits by inversion-consistent transfer learning. ACM
  Transactions on Graphics (TOG)  \textbf{40}(4),  1--13 (2021)

\bibitem{wang2009faceCUFS}
Wang, X., Tang, X.: Face photo-sketch synthesis and recognition. IEEE
  Transactions on Pattern Analysis and Machine Intelligence (PAMI)
  \textbf{31}(11),  1955--1967 (2009). \doi{10.1109/TPAMI.2008.222}

\bibitem{Wang2014SSIM}
Wang, Z., Bovik, A.C., Sheikh, H.R., Simoncelli, E.P.: Image quality
  assessment: from error visibility to structural similarity. IEEE Transactions
  on Image Processing  \textbf{13}(4),  600--612 (2004)

\bibitem{wang2019semi}
Xu, Z., Wang, H., Yang, Y.: Semi-supervised self-growing generative adversarial
  networks for image recognition. Multimedia Tools and Applications
  \textbf{80}(11),  17461--17486 (2021). \doi{10.1007/s11042-020-09602-1}

\bibitem{yi2019apdrawinggan}
Yi, R., Liu, Y.J., Lai, Y.K., Rosin, P.L.: Apdrawinggan: Generating artistic
  portrait drawings from face photos with hierarchical gans. In: Proceedings of
  the IEEE/CVF Conference on Computer Vision and Pattern Recognition (CVPR).
  pp. 10743--10752 (2019)

\bibitem{yu2019free}
Yu, J., Lin, Z., Yang, J., Shen, X., Lu, X., Huang, T.S.: Free-form image
  inpainting with gated convolution. In: Proceedings of the IEEE/CVF
  International Conference on Computer Vision (CVPR). pp. 4471--4480 (2019)

\bibitem{Zhu2017Unpaired}
Zhu, J.Y., Park, T., Isola, P., Efros, A.A.: Unpaired image-to-image
  translation using cycle-consistent adversarial networks. In: 2017 IEEE
  International Conference on Computer Vision (ICCV). pp. 2242--2251 (2017).
  \doi{10.1109/ICCV.2017.244}

\bibitem{zhuang2021few}
Zhuang, N., Yang, C.: Few-shot knowledge transfer for fine-grained cartoon face
  generation. In: 2021 IEEE International Conference on Multimedia and Expo
  (ICME). pp.~1--6. IEEE (2021)

\end{thebibliography}

\end{document}